\documentclass[11pt,a4paper]{article}
\usepackage[hyperref]{emnlp2018}
\usepackage{times}
\usepackage{latexsym}

\usepackage{amsmath}
\usepackage{bm}
\usepackage{color}
\usepackage{multirow}
\usepackage{subfig}
\usepackage{arydshln}
\usepackage{amsfonts}
\usepackage{graphicx} 
\usepackage{graphics}
\usepackage{CJKutf8}

\usepackage{balance}

\aclfinalcopy 

\newcommand*\mean[1]{\overline{#1}}

\title{Modeling Localness for Self-Attention Networks}

\author{Baosong Yang\\\normalsize University of Macau\\{\normalsize \tt nlp2ct.baosong@gmail.com} \And
Zhaopeng Tu\thanks{~Zhaopeng Tu and Derek F. Wong are the co-corresponding authors of the paper. This work was conducted when Baosong Yang was interning at Tencent AI Lab.}\\\normalsize Tencent AI Lab\\{\normalsize \tt zptu@tencent.com} \And
Derek F. Wong$^*$\\\normalsize University of Macau\\{\normalsize \tt derekfw@umac.mo} \AND
Fandong Meng\\\normalsize Tencent AI Lab\\{\normalsize \tt fandongmeng@tencent.com} \And
Lidia S. Chao\\\normalsize University of Macau\\{\normalsize \tt lidiasc@umac.mo} \And
Tong Zhang\\\normalsize Tencent AI Lab\\{\normalsize \tt bradymzhang@tencent.com}
}

\begin{document}
\maketitle
\begin{abstract}
Self-attention networks have proven to be of profound value for its strength of capturing global dependencies.
In this work, we propose to model localness for self-attention networks, which enhances the ability of capturing useful local context. We cast localness modeling as a learnable Gaussian bias, which indicates the central and scope of the local region to be paid more attention.
The bias is then incorporated into the original attention distribution to form a revised distribution.
To maintain the strength of capturing long distance dependencies 
and enhance the ability of capturing short-range dependencies, we only apply localness modeling to lower layers of self-attention networks.
Quantitative and qualitative analyses on Chinese$\Rightarrow$English and English$\Rightarrow$German translation tasks 
demonstrate the effectiveness and universality of the proposed approach.
\end{abstract}

\section{Introduction}

Recently, a new simple architecture, the \textsc{Transformer}~\cite{Vaswani:2017:NIPS}, that based solely on attention mechanisms has attracted increasing attention in machine translation community. 
Instead of using complex recurrent or convolutional neural networks, \textsc{Transformer} implements encoder and decoder as self-attention networks to draw global dependencies between input and output. 
By further parallel performing (multi-head) and stacking (multi-layer) attentive functions, 
\textsc{Transformer} has achieved state-of-the-art performance on various translation tasks \cite{shaw2018self,hassan2018achieving}.

One strong point of self-attention networks is the strength of capturing long-range dependencies by explicitly attending to all the signals. In this way, a representation is allowed to build a direct relation with another long-distance representation. Accordingly, it can serve as the role of RNN and CNN to capture both the short- and long-range relations among the representations.

Self-attention networks fully take into account all the signals with a weighted averaging operation. We argue that such operation disperses the distribution of attention, which results in overlooking the relation of neighboring signals. Recent works have shown that self-attention networks benefit from locality modeling. For example,~\newcite{shaw2018self} introduced relative position encoding to consider the relative distances between sequence elements, which produces substantial improvements on the translation task. 
\newcite{sperber2018self} modeled the local information by restricting self-attention model to neighboring representations, which boosts performance on long-sequence acoustic modeling. Although not for self-attention,~\newcite{luong2015effective} proposed a local attention model for translation task, which looks at only a subset of source words at a time. Inspired by these studies, we propose more flexible strategies for modeling localness for self-attention networks in this work.



Specifically, we cast the localness modeling as a learnable Gaussian bias, in which a central position (i.e. mean of the position) and a dynamic window (i.e. deviation of the distribution) are predicted with the intermediate representations in the self-attention network. Intuitively, the central position and the window respectively denote the center and the scope of the locality to be paid more attention. The learned Gaussian bias is then incorporated into the original attention distribution to form a revised distribution, which considers the expected local context.

Some researchers may doubt that self-attention networks augmented localness modeling focuses leanings toward local context, which weakens its strength of capturing long-range dependencies. Our extensive analyses can dispel such doubt by showing that the potential problem is compensated by multi-layer multi-head self-attention networks. First,  multi-head attention attends to local regions centered at different positions, which can constitute the complete information of an input sequence. Second, we found that self-attention models tend to capture short-range dependencies among neighboring words in lower layers, while capture long-range dependencies beyond phrase boundaries in higher layers. Accordingly, we only apply localness modeling to lower layers.

We conducted experiments on two widely-used WMT14 English$\Rightarrow$German and WMT17 Chinese$\Rightarrow$English translation tasks.
The proposed approach consistently improves translation performance over the strong \textsc{Transformer} baseline, demonstrating its effectiveness and universality. In addition, our approach is complementary to the relative position encoding~\cite{shaw2018self}, and combining them can further improve translation performance.

\section{Background}
Attention model has recently been a basic module of most deep learning models. The mechanism allows to dynamically select related representations as needed. 
In particular, it is very useful for generation models such as machine translation~\cite{bahdanau2015neural,luong2015effective,yang2017towards} and image captioning~\cite{xu2015show}.

\subsection{Self-Attention Model}

Recently, self-attention networks~\cite{Vaswani:2017:NIPS,shaw2018self,Shen:2018:AAAI} have attracted increasing attention 
due to their flexibility in parallel computation and dependency modeling. Self-attention networks calculate attention weights between each pair of tokens in a single sequence, thus can capture long-range dependency more directly than their RNN counterpart.


Formally, given an input sequence ${\bf x}=\{x_1, \dots, x_I\}$, each hidden state in the $l$-th layer is constructed by attending to the states in the $(l-1)$-th layer.\footnote{The first layer is the word embedding layer.}
Specifically, the $(l-1)$-th layer ${H}^{l-1} \in \mathbb{R}^{I\times d}$ is first transformed into the queries $Q \in \mathbb{R}^{I\times d}$, the keys $K \in \mathbb{R}^{I\times d}$, and the values $V \in \mathbb{R}^{I\times d}$ with three separate weight matrices. The $l$-th layer is calculated as:
\begin{eqnarray}
    H^l &=& \textsc{Att}(Q, K)\ V \label{eq:out},
\end{eqnarray}
where $\textsc{Att}(\cdot)$ is a dot-product attention model, defined as:
\begin{eqnarray}
    \textsc{Att}(Q, K) &=& softmax(energy) \label{eq:softmax} \\
    energy &=& \frac{QK^T}{\sqrt{d}}, \label{eq:sim}
\end{eqnarray}
where $\sqrt{d}$ is the scaling factor with $d$ being the dimensionality of layer states.

\subsection{Motivation}
The self-attention network models the global dependencies without regard to their distances, by directly attending to all the positions in an input sequence (i.e. Equation~\ref{eq:sim}).
We argue that self-attention can be further improved by taking into account the local context.
However, since the conventional self-attention models consider all of the words in a sequence, the weighted averaging inhibits the relation among the neighboring words.

\begin{figure*}[ht]
\begin{center}
\includegraphics[width=\textwidth]{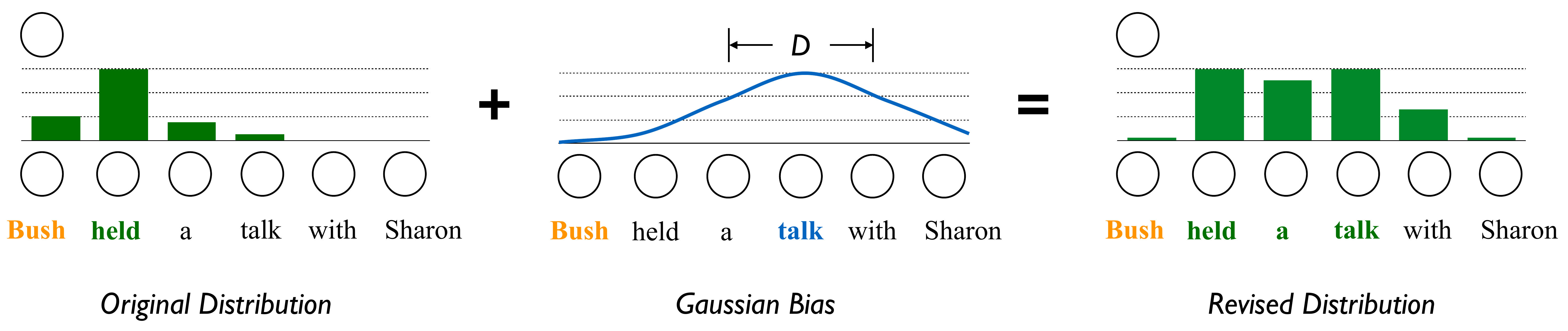}
\caption{
\label{fig:approach}
Illustration of the proposed approach. In this example, window size of 2 is used ($D=2$).
}
\end{center}
\end{figure*}

From a linguistic intuition, when a word $x_i$ is aligned to another word $x_j$, we also expect $x_i$ to align mainly to the neighboring words of $x_j$, so as to capture the phrasal patterns that contain useful local context information.
Take Figure~\ref{fig:approach} as an example, if ``{\em Bush}'' is aligned to ``{\em held}'' with high probability, we expect the self-attention model to pay more attention to the neighboring words ``{\em a talk}''. 
Consequently, the model is guided to capture the phrase ``{\em held a talk}''.

\section{Localness Modeling}

Figure~\ref{fig:approach} shows an example. We first learn a Gaussian bias, which is centered around the word ``{\em talk}'' (it is not necessary to be consistent with the original attention distribution), with a window size being 2 (in practice, it is a float number in our model). 
The distribution of attention is then regularized with the learned Gaussian bias to produce the final distribution, which pays more  
attention to the local context around the word ``{\em talk}''.

\subsection{\bf Localness Modeling as a Gaussian Bias}
\label{sec:gaussian}
Specifically, 
a  Gaussian bias $\emph{G}$ is placed to mask the logit similarity $energy$ in Equation~\ref{eq:softmax}, namely:
\begin{eqnarray}
    \textsc{Att}(Q, K) &=& softmax(energy + \emph{G}).  
\end{eqnarray}
The first term is the original dot product self-attention model.
$\emph{G}\in \mathbb{R}^{I\times I}$ is a favor alignment position matrix ($I$ denotes the sequence length). 
The element $\emph{G}_{i,j} \in [0,-\infty)$ measures the tightness between the word $x_j$ and the predicted central position $P_i$:
 \begin{equation}  
  \emph{G}_{i,j} = -\frac{(j-P_i)^2}{2{\sigma_i}^2},
 \end{equation}  
where $\sigma_i$ denotes the standard deviation which is empirically set as $\sigma_i = \frac{D_i}{2}$, and $D_i$ is a window size. Note that, due to the exponential operation in $softmax$ function, adding the logit similarity $energy$ with a bias $\in [0,-\infty)$ approximates to multiplying the attention distribution by a weight $\in [1,0)$. The position and window size can be calculated as:
\begin{equation}
 \begin{bmatrix}
 P_i\\ 
D_i
\end{bmatrix} = I \cdot \text{sigmoid}(\begin{bmatrix}
 p_i\\ 
z_i
\end{bmatrix}).
\label{eqn:gaussian}
\end{equation}
The scalar factor $I$ is used to regulate $P_i$ and $D_i$ to real value numbers between $0$ and the length of input sequence. The predictions are conditioned on two scalar $p_i$ and $z_i$ respectively. 

\subsection{\bf Central Position Prediction}

  Since the prediction of each central position depends on its corresponding query vector,\footnote{For the input of feed-forward network, we also tried an additive term to consider the weighted context $O_i$ (Equation~\ref{eq:out}), namely: $\tanh (W_p Q_i + W_o O_i)$. Our experimental results showed that there is no progressive improvement.} we simply apply a feed-forward network to transform $Q_i$ into a positional hidden state, which is then mapped into the scalar $p_i$ by a linear projection $U_p\in\mathbb{R}^{d}$, namely: 
 \begin{equation}
  p_i = {U_p}^T \tanh (W_p Q_i),
  \label{eqn:position}
 \end{equation}
where $W_p\in \mathbb{R}^{d \times d}$ is the model parameter. 

\subsection{Window Size Prediction}
\label{sec:window}
 Several alternative strategies are proposed to select the window size. Except 
 a non-parametric approach, the other two define parametric windows. Among the parametric methods, the first strategy assigns a unified window size to all the hidden states in a layer, so as to consider the context of the sequence, while the second one calculates a distinct window size for each hidden state.
\paragraph{\bf Fixed-Window}

A simple choice is to use a pre-defined window size $D$, which is a constant number throughout the whole training and testing process. In this study, following the common practice \cite{luong2015effective}, $D$ is set to 10.

\paragraph{\bf Layer-Specific Window} 
  
Furthermore, an interpretable way to select the window size is to account for the context of the sequence by summarizing the information from all the representations in a layer. In this study, we assign the mean of keys $\mean{\bf K}$ to represent the semantic context. Thus, the unified scalar $z$ of a layer  is defined as:

    \begin{equation}
     z = {U_d^T\tanh (W_d \mean{\bf K})},
  \end{equation} 
  where $W_d\in \mathbb{R}^{d \times d}$ and $U_d\in\mathbb{R}^{d}$ are learnable parameters.

\paragraph{\bf Query-Specific Window} 
The last strategy provides a more flexible manner to differentiate the scope by conditioning on each query. Similar to the prediction of the central position (Equation~\ref{eqn:position}), the query-specific window can be formally expressed as:
 \begin{equation}
  z_i = {U_d}^T \tanh (W_p Q_i).
  \label{eqn:window}
 \end{equation}
Here, $U_d\in\mathbb{R}^{d}$ is a trainable linear projection. Note that, Equations~\ref{eqn:position} and \ref{eqn:window} share same parameter $W_p$ but use different $U_p$ and $U_d$.  The intuition behind this design is that the central position and window size interdependently locate the local scope, hence condition on the same hidden state. The distinct linear projections $U_p$ and $U_d$ are sufficient in distinguishing the two scalars, resulting in a smaller parameter size and faster computational speed than that of the layer-specific model.

\subsection{Incorporating into \textsc{Transformer}}
\label{sec:trans}

We evaluate our model on the advanced \textsc{Transformer} model~\cite{Vaswani:2017:NIPS}, which builds an encoder-decoder framework merely using attention networks.
Both the encoder and decoder are composed of a stack of $L=6$ layers, each of which has two sub-layers.
The first is a multi-head self-attention layer, and the second is a position-wise fully connected feed-forward layer. In this section, we describe how to apply our approach to \textsc{Transformer} by adapting to {\em multi-head} and {\em multi-layer} self-attention networks.

\paragraph{Adapting to Multi-Head Self-Attention}
Instead of performing a single attention function, the multi-head mechanism employs $M$ separate attention models with distinct parameters to jointly attend to the information from different representation subspaces at different positions.
Accordingly, we assign a distinct Gaussian bias to each attention head, and rewrite Equation~\ref{eqn:gaussian} as:
\begin{equation}
 \begin{bmatrix}
 P^m_i\\ 
D^m_i
\end{bmatrix} = I \cdot \text{sigmoid}(\begin{bmatrix}
 p^m_i\\ 
z^m_i
\end{bmatrix}),
\label{eqn:gaussian-multi-head}
\end{equation}
where $p^m_i$ and $z^m_i$ are trained with distinct parameters to predict the central position and window size for the $m$-th attention head. 

We argue that multi-head self-attention may benefit more from localness modeling. Multi-head attention captures different features by attending to different positions, which complements the localness modeling that may potentially ignore the global information. Experimental results in Table~\ref{tab:head} confirm our hypothesis by showing that localness modeling achieves more significant improvement when working with multi-head attention than its single-head counterpart.

\paragraph{Adapting to Multi-Layer Self-Attention}
Recent work shows that different layers capture different types of features.
~\newcite{Anastasopoulos:2018:NAACL} indicated that higher-level layers are more representative than lower-level layers, while~\newcite{Peters:2018:NAACL} showed that higher-level layers capture context-dependent aspects of word meaning while lower-level layers model aspects of syntax. 
One question naturally arises: {\em is it necessary to model localness for all layers?}

In this work, we investigate which levels of layers benefit most from the localness modeling. In addition, we visualize the Gaussian biases across layers, to better understand the behaviors of different attentive layers.

\section{Experiments}

\subsection{Setup}
\label{sec:setup}

To compare with the results reported by previous work~\cite{pmlr-v70-gehring17a,Vaswani:2017:NIPS,hassan2018achieving}, we conducted experiments on both Chinese$\Rightarrow$English (Zh$\Rightarrow$En) and English$\Rightarrow$German (En$\Rightarrow$De) translation tasks.
For the Zh$\Rightarrow$En task, the models were trained using all of the available parallel corpus from WMT17 dataset with maximum length limited to 50, consisting of about $20.62$ million sentence pairs. We used newsdev2017 as the development set and newstest2017 as the test set. 
For the En$\Rightarrow$De task, we trained on the widely-used WMT14 dataset consisting of about $4.56$ million sentence pairs.  The models were validated on  newstest2013 and examined on newstest2014. 
The Chinese sentences were segmented by the word segmentation toolkit {\em Jieba,}\footnote{\url{https://github.com/fxshy/jieba}} and the English and German sentences were tokenized using the scripts provided in Moses. 
Then, all tokenized sentences were processed by byte-pair encoding (BPE) to alleviate the Out-of-Vocabulary problem \cite{sennrich2015neural} with 32K merge operations for both language pairs.
The 4-gram NIST BLEU score \cite{papineni2002bleu} is used as the evaluation metric.

We evaluated the proposed approaches on advanced \textsc{Transformer} model~\cite{Vaswani:2017:NIPS}, and implemented on top of an open-source toolkit -- THUMT\footnote{\url{https://github.com/thumt/THUMT}}~\cite{zhang2017thumt}. We followed~\newcite{Vaswani:2017:NIPS} to set the configurations and reproduced their reported results on the En$\Rightarrow$De task.
We tested both the \emph{Base} and \emph{Big} models, which differ at the layer size (512 vs. 1024) and the number of attention heads (8 vs. 16). All the models were trained on eight NVIDIA P40 GPUs, each of which is allocated a batch of 4096 tokens.  In consideration of the computation cost, we studied the variations of the \emph{Base} model on Zh$\Rightarrow$En task, and evaluated the overall performance with the \emph{Big} model on both Zh$\Rightarrow$En and En$\Rightarrow$De translation tasks.

\subsection{Ablation Study}
\label{sec:ablation}

In the first series of experiments, we evaluated the impact of different components on the Zh$\Rightarrow$En validation set using the \textsc{Transformer-Base}.
First, we investigated the effect of different strategies to predict the localness window.
Then, we examined whether it is necessary to apply localness modeling to all the layers.
Finally, given that \textsc{Transformer} consists of encoder and decoder side self-attention as well as encoder-decoder attention networks, we checked which types of attention networks benefit most from the localness modeling.
To eliminate the influence of control variables, we conducted the first two ablation studies on encoder-side self-attention networks only. 

\begin{table}[t]
\centering
\renewcommand{\arraystretch}{1.2}
\begin{tabular}{l||c|cc}
    \bf Model & \bf Speed &  \bf  Dev & $\bigtriangleup$ \\ 
    \hline \hline
    Baseline    &   1.20 & 22.59  & -       \\ 
    \hline
    Fixed       &   1.14 & 23.07    &   + 0.48 \\ 
    Layer-Spec. &   1.07 & 23.13    &   + 0.54 \\
    Query-Spec. &   1.11 & 23.13    &   + 0.54 \\
  \end{tabular}
  \caption{Evaluation of various window prediction strategies for localness modeling, which is only applied to encoder-side self-attention network. ``Speed'' denotes training speed measured in steps per second.}
  \label{tab:scope}
\end{table}

\paragraph{Window Prediction Strategies}

As shown in Table~\ref{tab:scope}, 
all the proposed window prediction strategies consistently improve the model performance over the baseline, validating the importance of localness modeling in self-attention networks.
Among them, layer-specific and query-specific window outperform\footnote{{Although the differences are not always significant, the flexible strategy consistently outperforms its fixed counterpart across language pairs. For example, the query-specific strategy improves performance over the fixed-window model by +0.07 and +0.23 BLEU points on Zh-En and En-De validation sets, respectively.}} their fixed counterpart, showing the benefit that flexible mechanism is able to capture varying local context according to layer and query information.
Moreover, the flexible strategy does not reply on the hand-crafted parameters (e.g. the pre-defined window size), which makes model robustly applicable to other language pairs and NLP tasks.
Considering the training speed, we use the query-specific prediction mechanism as the default setting in subsequent experiments.

\begin{table}[t]
\centering
\begin{tabular}{c|c||c|cc}
\bf \# &  \bf Layers &    \bf Speed   &  \bf Dev   &  $\bigtriangleup$ \\
\hline \hline
1  & \text{[1-6]}   &  1.11    & 23.13 & - \\
\hline
2  & \text{[1-1]}   & 1.18   & 23.20 & + 0.07\\
3  & \text{[1-2]}   & 1.17   &  23.23 & + 0.10  \\
4  & \text{[1-3]}   & 1.15   & 23.29 & + 0.16 \\
5  & \text{[1-4]}   & 1.14   & 23.26  & + 0.13 \\
\hline
6  & \text{[4-6]}   & 1.15   & 23.22 & + 0.09 \\
\end{tabular}
\caption{Evaluation of different layers in the encoder, which are implemented as self-attention with query-specific localness modeling.}
\label{tab:layers}
\end{table}

\paragraph{Layers to be Applied}

In this experiment, we investigated the question of which layers should be applied with the localness modeling. Recent works show that different layers tend to capture different features, thus there may have different needs for the local context. We applied localness modeling to different combinations of layers, as shown in Table~\ref{tab:layers}. 
Clearly, modeling the localness for part of the layers consistently outperforms all layers in terms of  the training speed and translation quality, which again validates our claim.

Interestingly, the performance generally goes up with the increase of layers from bottom to top (Rows 2-4), while the trend does not hold when reaching the 4th-layer (Row 5).
In addition, the lower three layers benefit more from the localness modeling than that of the higher three layers (Rows 4 and 6).
These results reveal that lower-level layers benefit more from the local context.
Accordingly, we only model the localness in the lower three layers in the following experiments.

\paragraph{Attention Networks to be Applied}

\begin{table}[t]
  \centering
  \begin{tabular}{c|c|c||c|c}
   \bf ~~Enc~~~ &   \bf ~~Dec~~~   &   \bf Enc-Dec &    \bf Speed   &  \bf  Dev\\
    \hline \hline
    \checkmark  &   \texttimes   &  \texttimes & 1.15 & 23.29\\
    \checkmark  &	\checkmark   &  \texttimes & 1.10 & 23.27\\
    \checkmark  &	\texttimes   &  \checkmark & 1.08 & 23.33\\
    \checkmark   &  \checkmark   &  \checkmark & 1.02 & 23.19\\
  \end{tabular}
  \caption{Effect of localness modeling on different types of attention networks. ``Enc'' and ``Dec'' denote the encoder and decoder side self-attention networks respectively, while ``Enc-Dec'' represents the encoder-decoder attention network.}
  \label{tab:res}
\end{table}

\begin{table*}[t]
  \centering
  \begin{tabular}{l|l||rl|rl}
    \multirow{2}{*}{\bf System}  &   \multirow{2}{*}{\bf Architecture}  & \multicolumn{2}{c}{\bf Zh$\Rightarrow$En}  &  \multicolumn{2}{|c}{\bf En$\Rightarrow$De}\\
    \cline{3-6}
        &   &   \# Para.    &   BLEU    &   \# Para.    &   BLEU\\
    \hline \hline
    \multicolumn{6}{c}{{\em Existing NMT systems}} \\
    \hline
    \cite{wu2016google} &   \textsc{GNMT}             &  n/a &  n/a   &   n/a &   26.30\\ 
    \cite{pmlr-v70-gehring17a}  &   \textsc{ConvS2S}  &  n/a &  n/a   &   n/a &   26.36\\
    \hline
    \multirow{2}{*}{\cite{Vaswani:2017:NIPS}} &   \textsc{Transformer-Base}    &    n/a & n/a &  65M &   27.3\\ 
    &  \textsc{Transformer-Big}               &  n/a  &  n/a  &  213M &  28.4\\ 
    \hdashline
    \cite{hassan2018achieving}  &   \textsc{Transformer-Big}  &  n/a  &  24.2  &  n/a  & n/a\\
    \hline\hline
    \multicolumn{6}{c}{{\em Our NMT systems}}   \\ \hline
    \multirow{7}{*}{\em this work}  &   \textsc{Transformer-Base}  &    107.9M  & 24.13  &  88.0M  &   27.64\\
    &   ~~~ + Rel\_Pos \cite{shaw2018self}  & 108.0M  &   24.53 &  88.1M   &   27.94 \\  
    &   ~~~ + Localness                     & 108.7M  &   24.77$^\Uparrow$ &  88.8M   &   28.11$^\uparrow$ \\ 
    &   ~~~ + Localness + Rel\_Pos          & 108.8M  &   24.96$^\Uparrow$ &  88.9M   &   28.54$^\Uparrow$ \\ 
    \cline{2-6}
    &   \textsc{Transformer-Big}	                & 303.9M  &   24.56 &  264.1M &  28.58   \\ 
    &   ~~~ + Localness                     & 307.2M  &  25.03$^\uparrow$    & 267.4M  & 28.89 \\ 
    &   ~~~ + Localness + Rel\_Pos          & 307.3M  &  25.28$^\Uparrow$  & 267.5M & 29.18$^\Uparrow$ \\  
  \end{tabular}
  \caption{Comparing with the existing NMT systems on WMT17 Zh$\Rightarrow$En and WMT14 En$\Rightarrow$De test sets. ``\# Para." denotes the trainable parameter size of each model (M = million).  ``$\uparrow/\Uparrow$'': significant over the conventional self-attention counterpart ($p < 0.05/0.01$), tested by bootstrap resampling~\cite{Koehn2004Statistical}.} 
  \label{tab:exist}
\end{table*}

Table~\ref{tab:res} lists the results of localness modeling on different types of attention networks. 
As observed, modeling localness for decoder-side self-attention and encoder-decoder attention networks only marginally improves or even harms the translation quality.
We attribute the marginal improvement of the encoder-decoder attention network to the fact that it exploits the top-layer of encoder representations, which already embeds useful local context.
Concerning decoder-side self-attention network,~\newcite{zhang2018accelerating} pointed out that it tends to only focus on its nearby representation, which poses difficulties to modeling localness on the decoder side.
In the main experiments, we only applied localness modeling to the lower three layers of the encoder, which employs a query-specific window prediction strategy.


\subsection{Main Results}

In this section, we evaluated the proposed approach on both WMT17 Zh$\Rightarrow$En and WMT14 En$\Rightarrow$De translation tasks, as listed in Table~\ref{tab:exist}. 
Our baseline models, both \textsc{Transformer-Base} and \textsc{Transformer-Big}, outperform the reported results on the same data, which we believe make the evaluation convincing. As seen, modeling localness (``Localness'') consistently achieves improvement across language pairs and model variations, demonstrating the efficiency and universality of the proposed  approach.

We also re-implemented the relative position encoding (``Rel\_Pos'') that recently proposed  by~\newcite{shaw2018self}, which considers the relative distances between sequence elements.
{Both \newcite{shaw2018self} and our work have shown that explicitly modeling locality for self-attention networks can improve the model performance.} This indicates that it is necessary to enhance the locality modeling for Transformer. Besides, our approach is complementary to theirs, and combining them is able to further improve the translation performance. We attribute this to the fact that the two models modeling localness from two different aspects: 
{ First, the position embeddings are the same across different positions (if the absolute positions or relative positions are the same) and training examples, our model assigns a distinct localness bias to each position from layer to layer. Second, contrast to position encoding which learns the locality through the positional information in embeddings, our model directly revises the attention distribution to focus on a local space. 
}

\section{Analysis}

We conducted extensive analyses to better understand our model in terms of its compatibility with multi-head and multi-layer attention networks, as well as building the ability of capturing phrasal patterns. 
All the results are reported on Zh$\Rightarrow$En development set with \textsc{Transformer-Base}, unless otherwise stated.

\subsection{Compatibility with Multi-Head Attention}
\label{sec:multi-head}

In this section, we investigated whether multi-head attention and localness modeling are compatible from two perspectives: (1) whether multi-head attention benefits more from the localness modeling than its single-head counterpart; and (2) how does multi-head attention work together with localness modeling?


\begin{table}[h]
  \centering
  \setcounter{table}{4}
  \begin{tabular}{c||cc|cc}
\multirow{2}{*}{\bf Model} & \multicolumn{2}{c}{\bf 1-Head}  & \multicolumn{2}{|c}{\bf 8-Head}  \\ 
\cline{2-5}
    &   Dev &   $\bigtriangleup$    &   Dev &   $\bigtriangleup$\\
\hline \hline
 \textsc{Base}   & 22.05  & -       &   22.59   & - \\
 \textsc{Ours}   & 22.18  & + 0.13    &   23.29   & + 0.70\\
  \end{tabular}
  \caption{Evaluation of localness modeling on top of single and multiple attention heads.}
  \label{tab:head}
\end{table}

\paragraph{Multi-Head vs. Single-Head}
The single-head attention and multi-head attention differ at: the former uses a single 512-dimension attention head while the latter uses eight 64-dimension heads.
The results in Table~\ref{tab:head} confirm our claim by showing that multi-head attention indeed benefits more from our model than the single-head model (+0.70 vs. +0.13).
It should be noted that our model marginally improves the performance under single-head setting. One possible reason is that our model focuses more on local context thus may ignore global information, which cannot be complemented by the single-head attention.

\begin{figure}[h]
\begin{center}
\includegraphics[scale=0.75]{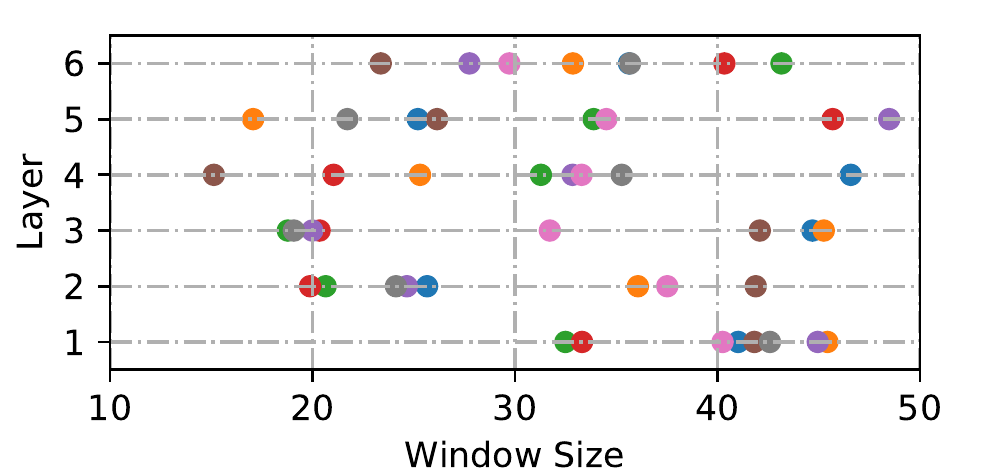}
\caption{Instructions of the learned window size by head-specific parametric model, where colors distinguish the heads. 
\label{fig:glob}
}
\end{center}
\end{figure}

\begin{figure*}[t]
\begin{center}
\includegraphics[scale=0.76]{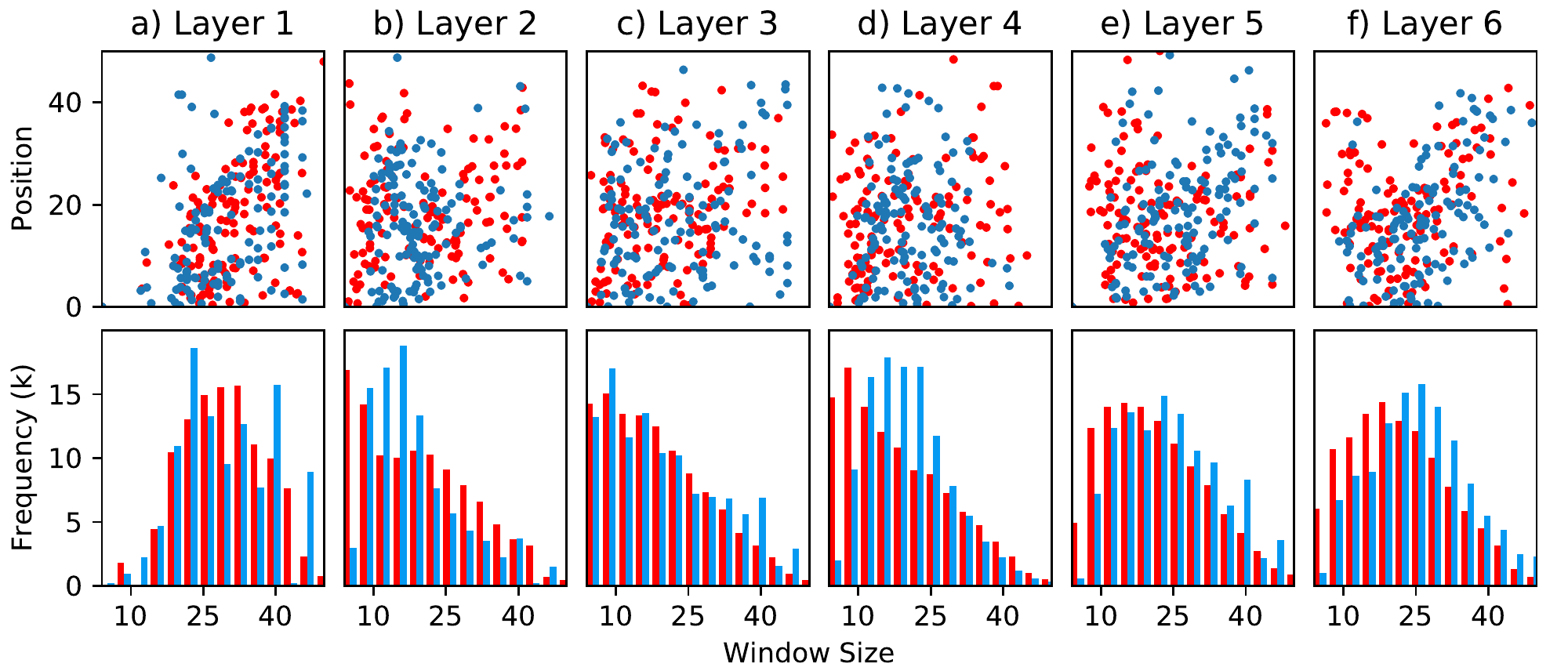}
\caption{
\label{fig:window_size}
Distribution of the local scopes learned by each attentive layer. The upper figures illustrate the distribution of the predicted pairs of central position (Y-axis) and its correspond window size (X-axis) in each layer, the samples are randomly selected from the development set. The lower figures show the distribution of the window size in each layer. Blue color represents the \textcolor{blue}{layer-specific parametric} approach, while the \textcolor{red}{query-specific parametric} method is indicated in red. 
}
\end{center}
\end{figure*}
\paragraph{Can Multi-Head Separate Locality?}

To simplistically visualize how heads cooperate in modeling localness, we propose an additional parametric model which is assigned a learnable but unified window size for each head, namely {\bf head-specific}. As a result, the window size $D^m$ of the $m$-th head is calculated as:
    \begin{equation}
     D^m = N\cdot \text{sigmoid}(z^m),
    \end{equation} 
where the scalar $z^m$ is a trainable parameter, $N=50$ denotes a pre-defined constant number. 

Figure~\ref{fig:glob} visualizes the distribution of the learned window size of each head, 
verifying that multi-head attention is able to capture diverse information by selecting suitable window sizes for different heads.
For example, in the middle-level layers, heads are assigned to consider both the global and local information by regulating the different window sizes. 
One interesting finding is that the distributions of window size are not exactly same in different layers, which is explored in more details in the next section.

\subsection{Analysis on Multi-Layer Attention}
\label{sec:multilayer}

In this section, we try to answer how does each layer learn the localness. We first investigated how the window size varies across layers. Then we checked the specific behavior of the first word embedding layer, which is inconsistent with the trend of other layers.

\paragraph{The Higher Layer, The Larger Scope}  \newcite{shi2016does} and \newcite{Vaswani:2017:NIPS} have shown that different layers have the abilities to distinguish and capture diverse syntactic context (e.g. the dependents between words). Figure~\ref{fig:window_size} shows 
the distribution of local scopes predicted by each layer. Except the first layer, the higher layers are more likely to pay attention to larger scopes, indicating that self-attention models tend to capture short-term dependencies among neighboring words in lower layers, while capture long-range dependencies beyond phrase boundaries in higher layers.


\paragraph{The Special First Layer} 
Inconsistent with the intuition which the lower layers may focus on local information, in common, the first layer is assigned with large scopes of local context. The same phenomenon has also occurred for head-specific model (Figure~\ref{fig:glob}).
Since the first layer represents word embeddings that are deficient in context, we argue that the self-attention model at first layer has to encode the representations with global context. In addition, experimental results in Table~\ref{tab:layers} (Row 2) show that despite its large local size, modeling localness at the first layer is still valid. 

 \begin{figure}[t]
\begin{center}
\includegraphics[width=0.45\textwidth]{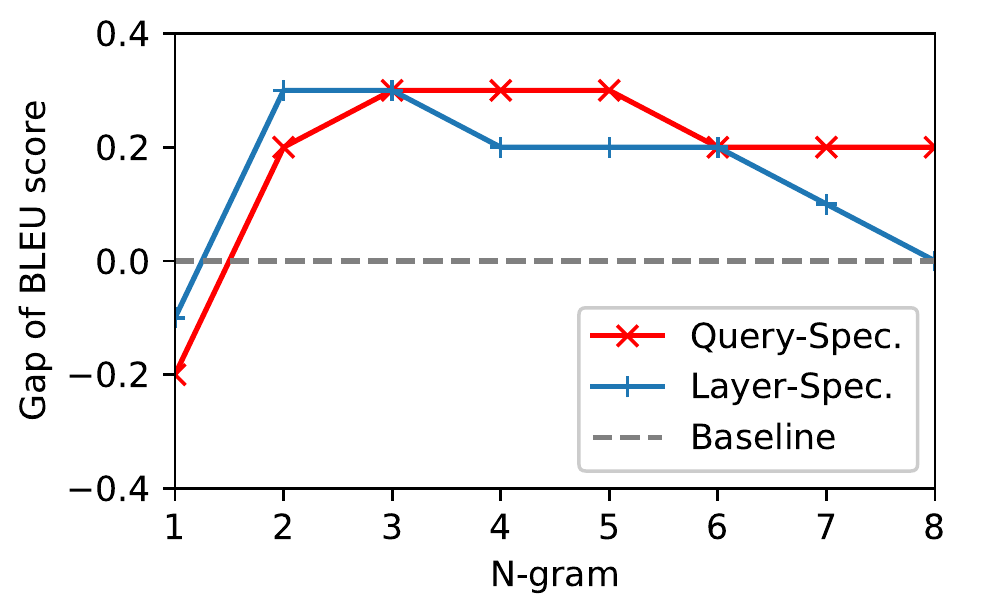}
\caption{Performance improvement according to N-gram. Y-axis denotes the gap of BLEU score between our model and baseline.}
\label{fig:ana}
\end{center}
\end{figure}

\subsection{Analysis on Phrasal Pattern}
\label{sec:pattern}

As aforementioned, one intuition of our approach is to capture useful phrase patterns. To evaluate the accuracy of phrase translations, we calculate the improvement of the proposed approaches over multiple N-grams, as shown in Figure~\ref{fig:ana}.
Although our models underperform the baseline on unigram translations, they consistently outperform the baseline on larger granularities, indicating that modeling locality can raise the ability of self-attention model on capturing the phrasal information.
Concerning the two variations, query-specific localness modeling surpasses its layer-specific counterpart on large phrases (i.g. 4-grams to 8-grams). We attribute this to the more modeling flexibility of query-specific strategy to differentiate the scope by conditioning on each query.



\section{Related Work}


A successful extension of neural language model is attention mechanism, which can directly capture long-distance dependencies by attending to previously generated words.~\newcite{Daniluk:2017:ICLR} proposed a {\em key}-{\em value}-{\em predict} attention to separate the key addressing, value reading, and word predicting functions explicitly.
\newcite{im2017distance} and \newcite{sperber2018self} adopted self-attention networks for acoustic modeling and natural language inference tasks, respectively.

\newcite{Vaswani:2017:NIPS} applied the idea of self-attention to neural machine translation. \newcite{Shen:2018:AAAI} and \newcite{Shen:2018:ICLR} proposed to improve the self-attention model with directional masks and multi-dimensional features.
Although the standard self-attention model can give more bias toward localness,\footnote{As pointed out by one reviewer, in the original self-attention model, there are some considerations about given more bias toward the localness. For example, base on the definition of the positional embeddings, the adjacent words will have more similar positional embeddings compared with more further words. After summing word embeddings and corresponding positional embeddings together, the model would prefer the local words.}
several studies show that explicitly modeling localness for self-attention model can further improve performance.
For example, \newcite{sperber2018self} showed that restricting the self-attention model on the neighboring representations performs better for longer sequences in acoustic modeling and natural language inference tasks.
Closely related to this work,~\newcite{shaw2018self} introduced relative position encoding to consider the relative distances between sequence elements. While they modeled localness from {\em static position embedding}, we improve locality modeling from {\em dynamically revising attention distribution}. Experimental results show that the two models are complementary to each other, and combining them can further improve performance.

Several researches have shown that explicitly modeling phrases is useful for neural machine translation~\cite{wang2017phrases,huang2018phrasal}. Our results confirm these findings.
Concerning attention models,~\newcite{luong2015effective} proposed a modification to look at only a subset of input words at a time. This can be regarded as a ``hard'' variation of our fixed-window strategy. In this study, we propose more flexible strategies for placing and zooming the local scope, which yield better results than the fixed scope.

\section{Conclusion}

In this work, we enhanced the ability of capturing local context for self-attention networks with a learnable Gaussian bias. We proposed several strategies to learn the scope of the local context, and found that a query-specific mechanism yielded the best result due to its more modeling flexibility.
Experimental results on widely-used English$\Rightarrow$German and Chinese$\Rightarrow$English translation tasks demonstrate the effectiveness and universality of the proposed approach.
By visualizing the scopes of the learned Gaussian biases, we found that the higher the layer, the larger scope the bias, which is consistent with the findings in previous work~\cite{shi2016does,Peters:2018:NAACL}.

As our approach is not limited to specific tasks, it is interesting to validate our model in other tasks, such as reading comprehension, language inference, and stance classification~\cite{Xu:2018:ACL}.
Another promising direction is to design 
more powerful localness modeling techniques, such as incorporating linguistic knowledge (e.g. phrases and syntactic categories).
It is also interesting to combine with other techniques~\cite{shaw2018self,Shen:2018:AAAI,Dou:2018:EMNLP,Li:2018:EMNLP} to further improve the performance of Transformer.

\section*{Acknowledgments}
This work was supported in part by the National Natural Science Foundation of China (Grant No. 61672555), the Joint Project of Macao Science and Technology Development Fund and National Natural Science Foundation of China (Grant No. 045/2017/AFJ) and the Multiyear Research Grant from the University of Macau (Grant No. MYRG2017-00087-FST).
We thank the anonymous reviewers for their insightful comments.

\balance
\bibliography{emnlp2018}
\bibliographystyle{acl_natbib_nourl}

\end{document}